\documentclass[sigconf]{acmart}
\usepackage{graphicx}
\usepackage{subfigure}
\usepackage{hyperref}
\usepackage{balance}
\AtBeginDocument{%
  \providecommand\BibTeX{{%
    \normalfont B\kern-0.5em{\scshape i\kern-0.25em b}\kern-0.8em\TeX}}}

\copyrightyear{2020} 
\acmYear{2020} 
\setcopyright{acmcopyright}\acmConference[MM '20]{Proceedings of the 28th ACM International Conference on Multimedia}{October 12--16, 2020}{Seattle, WA, USA}
\acmBooktitle{Proceedings of the 28th ACM International Conference on Multimedia (MM '20), October 12--16, 2020, Seattle, WA, USA}
\acmPrice{15.00}
\acmDOI{10.1145/3394171.3413548}
\acmISBN{978-1-4503-7988-5/20/10}

\begin{document}
\fancyhead{}

\title{MS$^2$L: Multi-Task Self-Supervised Learning \\ for Skeleton Based Action Recognition}

\author{Lilang Lin}
\affiliation{%
  \institution{Wangxuan Institute of Computer Technology, \\Peking University}}
\email{linlilang@pku.edu.cn}

\author{Sijie Song}
\affiliation{%
  \institution{Wangxuan Institute of Computer Technology, \\Peking University}}
\email{ssj940920@pku.edu.cn}

\author{Wenhan Yang}
\affiliation{%
  \institution{Wangxuan Institute of Computer Technology, \\Peking University}}
\email{yangwenhan@pku.edu.cn}

\author{Jiaying Liu}
\authornotemark[0]
\authornote{Corresponding author. This work was supported by the National Key R$\&$D Program of China under Grand No.2018AAA0102700, National Natural Science Foundation of China under contract No.61772043, and Beijing Natural Science Foundation under contract No.4192025.}
\affiliation{%
  \institution{Wangxuan Institute of Computer Technology, \\Peking University}}
\email{liujiaying@pku.edu.cn}


\begin{abstract}
  In this paper, we address self-supervised representation learning from human skeletons for action recognition. Previous methods, which usually learn feature presentations from a single reconstruction task, may come across the overfitting problem, and the features are not generalizable for action recognition. Instead, we propose to integrate multiple tasks to learn more general representations in a self-supervised manner. To realize this goal, we integrate motion prediction, jigsaw puzzle recognition, and contrastive learning to learn skeleton features from different aspects. Skeleton dynamics can be modeled through motion prediction by predicting the future sequence. And temporal patterns, which are critical for action recognition, are learned through solving jigsaw puzzles. We further regularize the feature space by contrastive learning. Besides, we explore different training strategies to utilize the knowledge from self-supervised tasks for action recognition. We evaluate our multi-task self-supervised learning approach with action classifiers trained under different configurations, including unsupervised, semi-supervised and fully-supervised settings. Our experiments on the NW-UCLA, NTU RGB+D, and PKUMMD datasets show remarkable performance for action recognition, demonstrating the superiority of our method in learning more discriminative and general features. Our project website is available at \url{https://langlandslin.github.io/projects/MSL/}.
\end{abstract}

\begin{CCSXML}
    <ccs2012>
       <concept>
           <concept_id>10010147.10010178.10010224</concept_id>
           <concept_desc>Computing methodologies~Computer vision</concept_desc>
           <concept_significance>500</concept_significance>
           </concept>
     </ccs2012>
\end{CCSXML}
    
\ccsdesc[500]{Computing methodologies~Computer vision}

\keywords{Action recognition, multi-task, self-supervised learning}


\maketitle

\begin{figure}[t]
  \centering
  \includegraphics[width=0.5\textwidth]{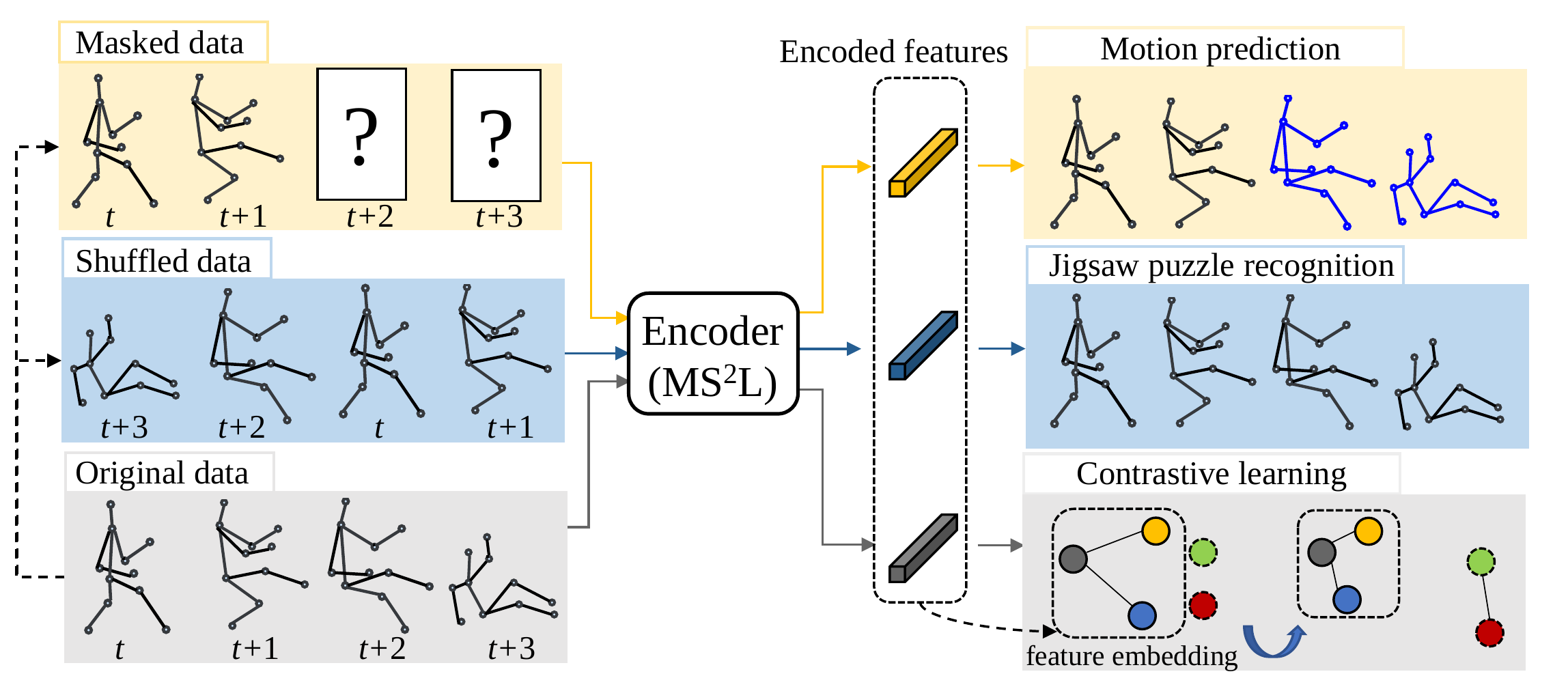}
  \caption{Our proposed approach (MS$^2$L) learns more general and discriminative features benefiting from multiple self-supervised tasks.}
  \label{fig:teaser}
\end{figure}

\section{Introduction}
Action recognition is a fundamental yet challenging problem in computer vision research. Demands on action recognition are growing rapidly to facilitate the applications such as video surveillance, human-computer interaction, video understanding. 

In the past few years, many works~\cite{hadsell2006dimensionality, wang2012mining, CVIU11SurveyAction} have been developed on recognition based on RGB videos and achieved many significant results. However, processing RGB videos can be very time-consuming and require a large storage space. Another data modality, human skeletons, which represent a person by the 3D coordinate positions of skeletal joints, draw much attention due to the light-weight representations, the robustness to variations of viewpoints, appearances, and surrounding distractions. Furthermore, skeleton sequences can be regarded as a high-level representation for human behavior, which attracts many researchers to study skeleton-based action recognition~\cite{zhu2015co,song2017end,du2016representation,zhang2017view,zhang2018adding,si2019attention,yan2018spatial}. Leveraging the merits of recurrent layers, many works~\cite{zhu2015co,song2017end,du2016representation,zhang2017view,zhang2018adding} build their framework based on Recurrent Neural Networks (RNN) to model temporal evolution of different actions. Considering that skeletons are naturally with graph structures, graph convolution networks (GCN) are applied in skeleton-based action recognition and show outstanding performance ~\cite{si2019attention,yan2018spatial}. However, these models are trained in a fully-supervised manner and require massive labeled training examples. Besides, annotating training data can be tedious and expensive. How to effectively learn feature representations from skeleton data with less annotation efforts remains a concerned problem.

%However, the video (RGB) datasets can occupy plenty of space and contain much irrelevant information about recognition, making it challenging to keep the network robust to various backgrounds and appearances. Another modality of data, human skeletal joints, can be much smaller and convenient for training and regarded as a high representation of actions. However, keypoints include much less information for recognition than other modalitys, so to obtain a better method to utilize the information in keypoints more comprehensively remains a widely concerned problem.

%Moreover, labeling the large scale of actions examples is even more challenging and may contain several errors, since the annotators may assign a wrong label to a sequence or the sequence may be unclear. Thus, recent works fix their attention more and more on extracting features more effectively from unlabeled data. Some models receive masked input sequences and learn to re-generate data \cite{zheng2018unsupervised}. And some use KNN for classification \cite{su2019predict}. However, this still remains difficulty achieving a comparable results with supervised learning. The learned representations lose much useful information and are not general enough for different tasks.

Recently, there are a few attempts~\cite{zheng2018unsupervised,su2019predict} exploring representation learning from unlabeled skeleton data. These models achieve feature learning by an encoder-decoder structure, the input of which is masked or original skeleton sequences, and the goal is to reconstruct skeleton sequences from the encoded features. We argue there are two potential issues in the previous work: (1) The skeleton reconstruction focuses more on detailed skeleton coordinates, ignoring the high-level spatio-temporal information which is critical for action recognition. (2) Learning from a single task could lead to overfitting for the specific task~\cite{ruder2017overview}. Therefore, the learned features from previous works may not be discriminative and general enough for recognizing skeleton sequences.

To address the aforementioned issues, we propose a novel self-supervised learning method by optimizing multiple task simultaneously. As shown in Figure~\ref{fig:teaser}, we focus on combining different tasks to make the representations more diverse and describe different aspects of information. In our paper, we design three tasks, \emph{i.e.}, the generation task for motion prediction, the classification task for solving jigsaw puzzles, and contrastive learning based on skeleton transformations. We aim to learn skeleton dynamics from motion prediction, model temporal evolution through solving jigsaw puzzles, and further regularize the feature space by contrastive learning. To fully utilize the knowledge learned from self-supervised learning tasks and facilitate action recognition, we explore different training strategies to train the action classifier. We provide comprehensive evaluation and analysis in our experiments to demonstrate the superiority of our proposed self-supervised learning approach. 

%In this paper, we propose a novel self-supervised learning method to learn more useful representations, which combines three self-supervised tasks extended on skeleton data and two novel learning methods which can stably improve the accuracy. For the self-supervised tasks, we focus on combining different tasks to make the representations more diverse and contain various dimensions of information. And also, the tasks should relate to recognition. Thus, we choose motion prediction, jigsaw puzzle and contrastive learning for the default tasks. They focus on different aspects of the data: motion prediction aims to re-generate the original inputs, the jigsaw puzzle puts more attention on the temporal feature while contrastive learning extracts the common features between different action samples. For the two learning methods, we first explore a novel pretrain strategy which find a balance between self-supervised tasks and supervised learning tasks to make the representations more meaningful. Then we explore to train the tasks jointly as multi-task learning to research whether the tasks can interact with each other and perform better than single training.

In summary, our contributions include the following aspects:
\begin{itemize}
  \item We propose a multi-task self-supervised learning framework for skeleton based action recognition. We aim to learn comprehensive and general feature representations, benefiting from motion prediction, jigsaw puzzle recognition and contrastive learning. 
  %Apply multiple self-supervised learning tasks on skeleton data for action recognition to force the network to extract more comprehensive and useful representations. We extend three tasks to skeleton and propose a novel contrastive learning loss to adapt flexible number of positive and negative samples.
  \item To transfer the knowledge learned from self-supervised learning, we explore different training strategies, \emph{i.e.}, moving pre-training and jointly training, towards better action recognition performance.
  %By designing different training strategies, we explore diverse method to enhance the action recognition performance with self-supervised tasks.
  \item Exhaustive experiments on different datasets validate the capacity of our representations learned by self-supervised tasks, which show superiority in action recognition under different configurations, including the unsupervised, semi supervised and fully-supervised learning settings, as well as transfer learning.
  %Ablation studies and analysis about our approach also provide some insights to combination of the mutli-task learning and self-supervised learning.
\end{itemize}

The rest of the paper is organized as follows: Sec.~2 reviews previous works on self-supervised learning and skeleton-based action recognition. Sec.~3 introduces our proposed self-supervised learning approach and training strategies in detail. We present our experiment results and analysis in Sec.~4. Concluding remarks are given in Sec.~5.

\section{Related Work}
In this section, we first introduce related work on self-supervised learning, and then give a brief review on skeleton-based action recognition.

\subsection{Self-Supervised Learning}
Self-supervised learning aims to learn feature representations from a huge amount of unlabelled data. It has been verified that self-supervised pre-training can help supervised learning~\cite{erhan2010does} and it has a variety of applications in a broad range of computer vision topics~\cite{jang2018grasp2vec,owens2018audio}. Self-supervised learning is usually achieved by pretext tasks, which utilize easy-to-obtain automatically generated supervision without human expensive annotation.

Many efforts have been devoted to designing pretext tasks to learn image representations from unlabelled image data~\cite{doersch2015unsupervised,noroozi2016unsupervised,noroozi2018boosting,zhang2016colorful,gidaris2018unsupervised,wei2019iterative}. Doersh \emph{et al.}~\cite{doersch2015unsupervised} proposed to train a convolutional neural network to reorder perturbed image patches. Following the idea, the works in~\cite{noroozi2016unsupervised,noroozi2018boosting,wei2019iterative} predict a permutation of multiple shuffled image patches to better model spatial relationships, which are called jigsaw puzzles. There are also other pretext tasks, such as colorizing grayscale images~\cite{zhang2016colorful} or predicting image rotation angles~\cite{gidaris2018unsupervised,zhai2019s4l}. More recently, Chen \emph{et al.}~\cite{chen2020simple} proposed a visual representation learning method with contrastive learning for contrastive prediction task, which forces on feature representation between positive pairs more similar than those between negative ones.

Recent studies also pay attention to representation learning for sequential data such as videos. A common way is to predict the video frame orders~\cite{misra2016shuffle,lee2017unsupervised,fernando2017self} to learn the temporal patterns. To further learn spatio-temporal representations, Vondrick \emph{et al.}~\cite{vondrick2016generating} proposed a spatio-temporal 3D convolution integrated with a generative adversarial network. Kim \emph{et al.}~\cite{kim2019self} tackled the problem by solving space-time cubic puzzles inspired by jigsaw puzzles in the image domain. A more recent work~\cite{wang2019self} improves the spatio-temporal feature representations in a finer granularity by regressing motion and appearance statics along spatial and temporal dimensions. In our work, we use multiple tasks to learn spatial and temporal patterns, respectively. Besides, we apply the contrastive learning to constrain the feature space by sampling positive and negative pairs.

\subsection{Skeleton-Based Action Recognition}
\begin{figure*}[h]
  \includegraphics[width=1.0\textwidth]{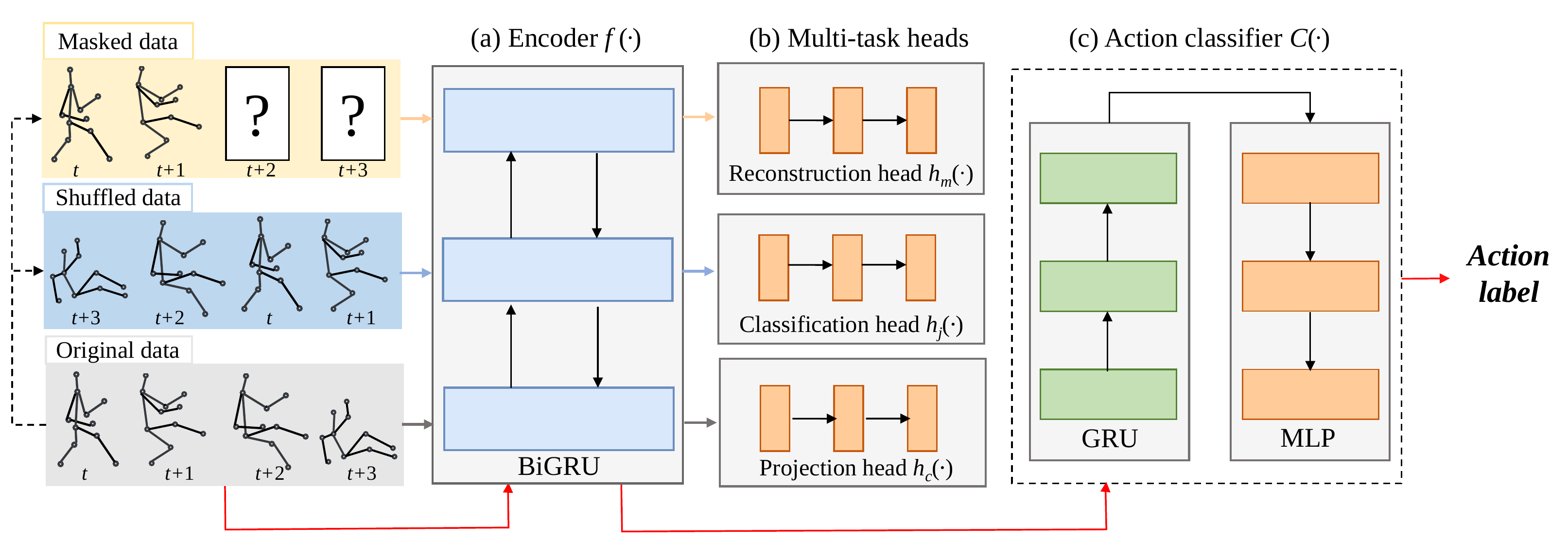}
  \caption{The structure of our network. (a) Encoder. (b) Multi-task heads. (c) Classifier for action recognition. We use yellow, blue and grey arrows to indicate the pipeline for motion prediction, jigsaw puzzle recognition and contrastive learning, respectively. Action recognition is achieved with the red pipeline.}
  \label{fig:pipeline}
\end{figure*}
%Prior works for skeleton-based action recognition are generally based on hand-crafted features. Recent works fix more attention on deep learning and neural networks. Du et al. \cite{du2015hierarchical} applies a hierarchical RNN to process body keypoints. Then an attention-based LSTM is proposed to automatically select important skeleton points and video frames(Song et al. \cite{song2017end}). Another work(Zhu et al. 2016 \cite{zhu2015co}) designed a model utilizing co-occurring joints as a strong discriminative feature and using a connection matrix to learn the mappings between co-occurrence of joints and the human action. Recently, graph convolution networks become popular in skeleton-based action recognition methods. Yan et al. \cite{yan2018spatial} proposes spatial Temporal Graph Convolution Networks to extract both the spatial and temporal features from skeleton data. Thus, two-stream GNN \cite{shi2019two} can be used for representation and learn the graph an adaptive manner.
Early skeleton-based action recognition methods are generally based on hand-crafted features by utilizing the geometry relationships of skeletal joints~\cite{vemulapalli2014human,vemulapalli2016rolling,wang2012mining,goutsu2015motion,lv2006recognition}. Recent methods for skeleton-based action recognition pay more attention to utilizing deep networks as their basic models. Benefiting from the merits of recurrent layers for sequential data, Du \emph{et al.}~\cite{CVPR15HRNN} proposed a pioneer work based on a hierarchical recurrent neural network. Zhu \emph{et al.}~\cite{zhu2015co} explored the co-occurrence of joints by introducing a group sparsity constraint on the recurrent neural network. To more adaptively learn the co-occurrence patterns of skeletal joints, attention-based methods are proposed to automatically select important skeletal joints~\cite{song2017end,zhang2018adding,song2018spatio} and video frames~\cite{song2017end,song2018spatio}. However, recurrent neural networks always suffer from gradient vanishing~\cite{hochreiter2001gradient}, which may cause optimization problems. Then, convolutional neural networks attract more attention for the skeleton-based action recognition. To adapt the input for convolutional neural network, a new representation for 3D skeleton data is proposed in~\cite{ke2017new}, which converts the problem of action recognition to image classification. To better model the structural information of the skeleton data, Yan \emph{et al.}~\cite{yan2018spatial} employed graph convolution networks to represent skeleton data on graphs. To make the graph representation more flexible, attention mechanisms are applied in~\cite{si2019attention,shi2019two} to adaptively capture discriminative features based on spatial configurations and temporal dynamics.

Though these models have achieved excellent performance on skeleton-based action recognition, they heavily rely on expensive annotation of action sequences. To get rid of action labeling, Zheng \emph{et al.}~\cite{zheng2018unsupervised} proposed an unsupervised framework using an encoder-decoder structure to re-generate the masked input sequence. To force the encoder to learn a more informative representation, a more recent work~\cite{su2019predict} is proposed to improve the encoder and weaken the decoder. However, the models in~\cite{zheng2018unsupervised,su2019predict} are learned from a single task, \emph{i.e.}, skeleton sequence reconstruction, leading to limited feature representation ability. To learn more intrinsic features, our work introduces multiple self-supervised tasks to further enhance feature representation learning without the help of human annotations.

%Our baseline structure is similar to the structure in Su et al. \cite{su2019predict}. We utilize the output of encoder as representation. However, we apply multi-tasks for self-supervised learning. And we concern about the performance on the tasks we combine. We choose various tasks to restrict the feature space strongly.

%However, a extended training strategy is required for larger scale multi-view and multi-subject datasets. And different from using the masked ground truth as an input into the decoder, Su \cite{su2019predict} improves the encoder and weakens the decoder.

\section{Multiple Self-Supervised Learning (MS$^2$L)}
In this section, we present our self-supervised learning techniques. We first provide a general description of our approach. Then, we introduce specific instantiations of our approach.

\subsection{Preliminaries}
We focus on the self-supervised feature learning for skeleton data. Then, we apply the learned features on skeleton-based action recognition. Basically, our overall framework consists of an encoder $f(\cdot)$ to extract features from skeleton data, and an action classifier $C(\cdot)$ to assign action labels to the input sequence. Supposing the $i^{th}$ skeleton sequence is $\mathbf{X}^i = \{\mathbf{x}^i_1, \dots, \mathbf{x}^i_T\}$, where $\mathbf{x}^i_t$ represents the $t^{th}$ frame. Then, the result for action recognition is $p^i = C\left(f\left(\mathbf{X}^i\right)\right)$, where $p^i$ is the probability distribution over all the action categories. In our work, our goal is to learn powerful feature representations from the encoder $f(\cdot)$ with self-supervised learning. Besides, we explore different settings and strategies to train the action classifier $C(\cdot)$ with the learned features.

\subsection{Multiple Self-Supervised Tasks}
We now describe our self-supervised learning techniques. To learn generalizable and robust skeleton features, we consider multiple self-supervised tasks, \emph{i.e.}, the generation task for motion prediction, the classification task for solving video jigsaw puzzles, and contrastive learning based on skeleton transformations. We aim to model skeleton dynamics through motion prediction and learn temporal patterns by solving jigsaw puzzles. Finally, we utilize contrastive learning to further regularize the feature space for more inherent representations. Figure~\ref{fig:pipeline} shows the pipeline of our model. The tasks share the encoder $f(\cdot)$ and adopt different heads for different objectives. Next, we present these self-supervised tasks in detail, respectively.

\begin{figure}[ht]
  \includegraphics[width=0.5\textwidth]{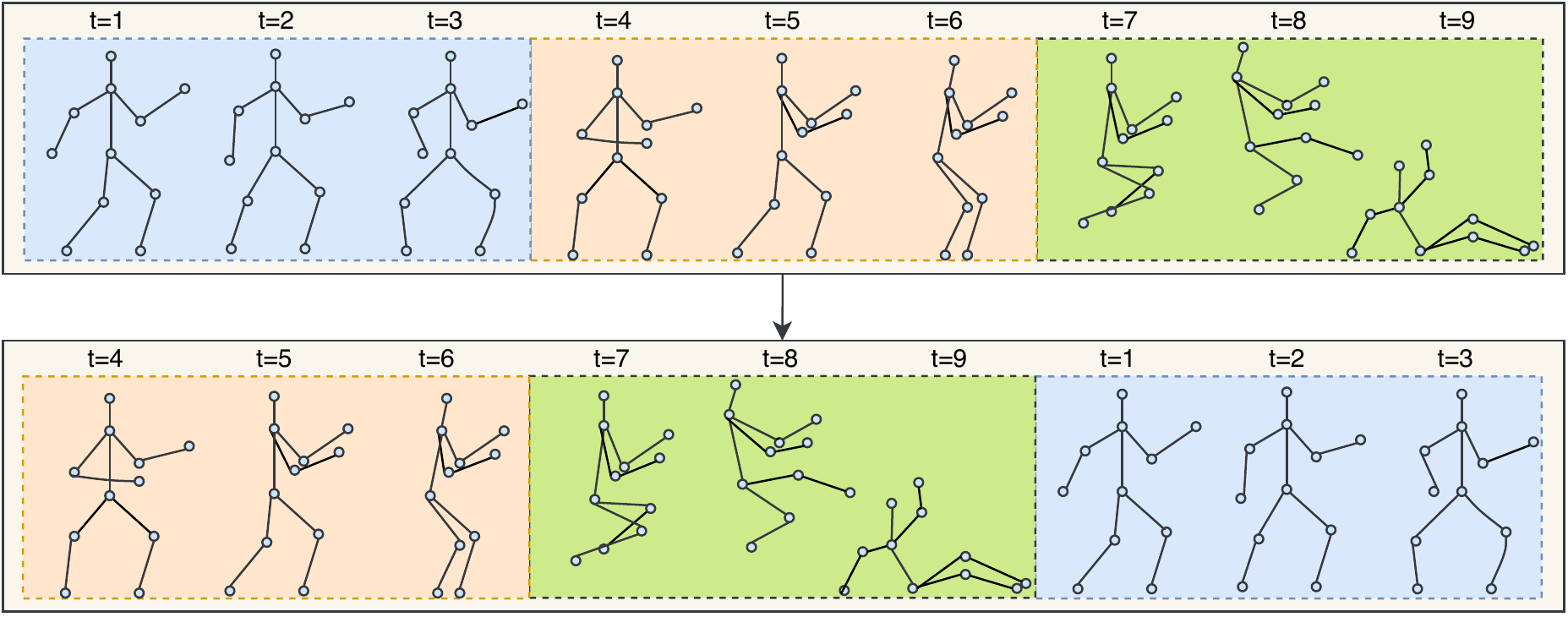}
  \caption{Our method of skeleton jigsaw puzzles. Different color means different segments and we shuffle these segments randomly to create various permutations.}
  \label{fig:jigsaw}
\end{figure}

\noindent\textbf{Motion Prediction.} Given the past motion sequence, the motion prediction task focuses on forecasting the most likely future poses of a person by modeling skeleton dynamics. Inspired by \textit{Seq2Seq}~\cite{sutskever2014sequence}, we apply an encoder-decoder with recurrent layers to achieve the task. 
%More specifically, our model consists of an encoder $f(\cdot)$ and a decoder $h_d(\cdot)$. The encoder $f(\cdot)$ is a bidirectional GRU which reads in parts of the input sequences and extracts representations from inputs.
The encoder $f(\cdot)$ reads in parts of the input sequences and extracts representations from inputs. The decoder $h_m(\cdot)$, which is shown as a reconstruction head in Figure~\ref{fig:pipeline}, receives the learned representations and generates sequences to reconstruct the whole input sequences. To avoid overfitting, we augment the original data by injecting random noises into the input sequences. Those random noises are sampled from a Gaussian distribution to avoid the network remember the input sequences.

To formulate motion prediction, recall that the original skeleton sequence is $\mathbf{X}^i = \{\mathbf{x}^i_1, \dots, \mathbf{x}^i_T\}$, and the masked sequence is $\mathbf{X}^i_m = \{\mathbf{x}^i_1, \dots, \mathbf{x}^i_{T'} | {T'} < T\}$. Then, we inject random noise into the input sequence $\mathbf{X}^i_m$ to get the noisy input sequence $\tilde{\mathbf{X}}^i_m$.
The future motion sequence is predicted by $\hat{\mathbf{X}}^i_m = h_m\left(f\left(\tilde{\mathbf{X}}^i_m\right)\right)$, where $\hat{\mathbf{X}}^i_m = \{\hat{\mathbf{x}}^i_{T'+1}, \dots, \hat{\mathbf{x}}^i_T\}$, we use mean square error (MSE) to estimate the parameters of the network as follows:
\begin{equation}
\label{equ:motion_prediction}
    \mathcal{L}_m =\sum_{i=1}^N \sum_{t=T'+1}^T \|\hat{\mathbf{x}}^i_t - \mathbf{x}^i_t \|^2_2,
\end{equation}
where $N$ is the batch size.

\noindent\textbf{Jigsaw Puzzle.} Solving the problem of jigsaw puzzles aims to predict the correct permutation from the shuffled sequences.
In our work, we apply jigsaw puzzles for skeleton sequences in the temporal domain so the network is able to learn temporal patterns.
To generate puzzles from the skeleton sequences, each sequence is divided into $P$ segments equally and there are $\frac{T}{P}$ frames in a segment. We shuffle these segments randomly and there are P! ways to shuffle them. The network is trained to predict the correct order of the shuffled segments. An example of jigsaw puzzle can be viewed in Figure~\ref{fig:jigsaw}.

With the shared encoder $f(\cdot)$, we apply a classification head $h_j(\cdot)$ to obtain the classification results to recognize video jigsaw puzzles. Specifically, we use an MLP as our classification head. The task is trained with the loss $\mathcal{L}_j$, which is formulated as cross entropy loss for classification as follows:
\begin{equation}
  \label{equ:jigsaw_puzzle}
      \mathcal{L}_{j} = -\sum_{i=1}^N y^i \log h_j(f(\mathbf{X}_j^i)),
  \end{equation}
where $\mathbf{X}_j^i$ is the shuffled sequence of original data $\mathbf{X}^i$ and $y^i$ is one-hot vector indicating the action label.

\noindent\textbf{Contrastive Learning.}
To further regularize the feature learning and encourage the network to learn inherent representations, we adopt contrastive learning by mapping the transformed data into a common feature space.
Inspired by \textit{SimCLR}~\cite{chen2020simple}, our network learns representations by maximizing cosine similarity between transformed modalities of the same original data. 
%Extended from recent contrastive learning algorithms~\cite{chen2020a}, our network explores more samples of positive data and negative data. 
For each original sample, we consider multiple transformations. Specifically, we randomly sample $N$ examples and apply $(M - 1)$ kinds of transformation operators to obtain $NM$ samples. Then for each original sample, we can construct (M-1) positive pairs with its transformed samples, and construct negative pairs with other samples. 

A projection head $h_c(\cdot)$ is designed to map the encoded sequences into the feature space. Let $\mathbf{z}_1, \mathbf{z}_2, \dots, \mathbf{z}_{NM}$ be the feature extracted from the output of encoder $f(\cdot)$, for any integer $k$ from $1$ to $N$, $\mathbf{z}_{(k-1)M+1}$ is the original data and the sequences from $\mathbf{z}_{(k-1)M+2}$ to $\mathbf{z}_{kM}$ are the transformed samples from the original sequence $\mathbf{z}_{(k-1)M+1}$.  $\mathbf{\bar{z}}_i = \frac{1}{M} \sum_{j=(i-1)M+1}^{iM}\mathbf{z}_j$ denotes the mean features of original and transformed data for $\mathbf{z}_{(i-1)M+1}$.
Similar to recent works~\cite{chen2020simple}, we use $sim(\mathbf{x}, \mathbf{y}) = \mathbf{x}^T\mathbf{y}/\|\mathbf{x}\|_2\|\mathbf{y}\|_2$ define the cosine similarity between $\mathbf{x}$ and $\mathbf{y}$. We define the loss function as follows:
\begin{equation}
\label{equ:contrastive_learning}
\mathcal{L}_c = -\sum_{i=1}^{MN} log \frac{exp(sim(\mathbf{z}_i, \mathbf{\bar{z}}_k))}{\sum_{j=1}^N exp(sim(\mathbf{z}_i, \mathbf{\bar{z}}_j))},
\end{equation}
where $k$ is $\lceil \frac{i}{M} \rceil$. This extension can adapt to an arbitrary number of transformation operators and gain a better constraint in the feature space. In practice, we adopt two transformation operators in our work. One is temporal masking, and the other is temporal jigsaw, which are actually the input of motion prediction and jigsaw puzzle recognition, respectively.

\subsection{Training for Action Recognition}
%In the following section, we propose two methods of combining supervised training with self-supervised learning. And for supervised learning, a classifier is added on top of the encoder $f(\cdot)$.
With the feature representations from the encoder $f(\cdot)$, we build an action classifier $C(\cdot)$ on top of the encoder to achieve action recognition. We consider different settings to train the classifiers, including the unsupervised setting, semi-supervised setting and fully-supervised setting. In the unsupervised setting, the encoder is trained only with self-supervised tasks introduced above, and then we train the action classifier independently by optimizing the cross-entropy loss with the encoder fixed. In the semi-supervised and fully-supervised settings, we are allowed to train the encoder and classifier jointly. Here, we explore two different training strategies for the semi-supervised and fully-supervised settings towards better performance for action recognition. 

\noindent\textbf{Moving  Pretraining  Strategy.}
The previous pretraining method~\cite{zheng2018unsupervised} initializes the encoder with learned weights and finetunes the whole network. However, that may cause severe destruction of the extracted features learned from self-supervised tasks. To address the issue, we adopt a novel pretraining scheme, using a linear regularization mechanism to adjust the weights between self-supervised tasks and the action recognition task. This can help stabilize the training process when switching between different tasks.

Specifically, let $\mathcal{L}_{cls}$ be a standard cross-entropy loss for action recognition, $\mathcal{L}_{self} = \mathcal{L}_m + \mathcal{L}_j + \mathcal{L}_c$ is the sum of self-supervised learning losses from motion prediction, jigsaw puzzle recognition and contrastive learning.
When we initialize the encoder with the weights trained by self-supervised tasks, our network would take several epochs to perform moving pretrained supervised learning, during which we train the network with self-supervised tasks and supervised learning tasks jointly with a changeable parameter to adjust the proportion of two tasks with a loss as follows:
\begin{equation}
\label{equ:moving_pretrain}
  \mathcal{L}_{moving} = \theta \mathcal{L}_{cls} + (1 - \theta) \mathcal{L}_{self},  
\end{equation}
where $\theta$ increases from $0$ to $1$ linearly and is fixed at $1$ finally.

\noindent\textbf{Jointly Training Strategy.}
Another alternative to train for action recognition is to optimize the networks jointly from scratch. Training data are fed into the encoder to extract features and then into their corresponding heads and the action classifier. The loss function can be defined as follows:
\begin{equation}
\label{equ:joint_train}
    \mathcal{L}_{joint} = \mathcal{L}_{cls} + \omega \mathcal{L}_{self},
\end{equation}
where $\omega$ is a non-negative scalar weight to balance the two terms. In practice, $\omega$ is set as $1$.
We will show the action recognition performance with different training strategies and give an analysis on that in the experiments. 

\section{Experiment Results}
For evaluation, we conduct our experiments on the following three datasets: the North-Western UCLA dataset~\cite{wang2014cross-view}, the NTU RGB+D dataset~\cite{shahroudy2016ntu}, and the PKUMMD dataset~\cite{10.1145/3365212}. Our goal is to evaluate whether our feature encoder $f(\cdot)$ trained with the proposed self-supervised learning approach can generate good feature representations for action recognition. Thus, we consider action classifiers trained under different settings (\emph{i.e.}, unsupervised, self-supervised, and fully supervised).  We also apply our approach to transfer learning. Finally, we give an ablation study to illustrate the effectiveness of each component in our work. 
\subsection{Dataset and Settings}
\noindent\textbf{North-Western UCLA (NW-UCLA)~\cite{wang2014cross-view}} This dataset is captured by Kinect v1 and contains 1494 videos in 10 action categories performed
by 10 subjects. Each body has 20 skeleton joints. There are three views of each action and we use the first two views for training and the third view for testing, which contains $1,018$ videos and 462 videos, respectively. 

\noindent\textbf{NTU RGB+D Dataset (NTU)~\cite{shahroudy2016ntu}} This is a large scale dataset including $56,578$ videos with 60 action labels and 25 joints for each body, including interactions with pairs and individual activities.
We test our method under the cross-subject protocol, that the training and testing are split by different subjects, leading to $40,091$ videos for training and $16,487$ videos for testing.

\noindent\textbf{PKU Multi-Modality Dataset (PKUMMD)~\cite{10.1145/3365212}} PKU-MMD is a new large scale benchmark for continuous multi-modality 3D human action understanding and covers a wide range of complex human activities with well annotated information. It contains almost $20,000$ action instances and $5.4$ million frames in $52$ action categories. Each sample consists of $25$ body joints. PKUMMD consists of two subsets, \emph{i.e.}, part I and part II. Part I is an easier version for action recognition, while part II is more challenging with more skeleton noise caused by the large view variation. We conduct experiments under the cross subject protocol on the two subsets, respectively.

To train the network, all the skeleton sequences are temporally down-sampled to $200$ frames. For the motion prediction, we add random noise to
the former $50$ frames and mask the latter $150$ frames. For the skeleton jigsaw task, we divide the sequence into $3$ segments so there are $6$ ways
to shuffle the sub-sequences.

The architecture is set as four parts. First, the shared encoder $f(\cdot)$
is a $1$-layer bidirectional GRU with $30$ units in each layer. The reconstruction head $h_m(\cdot)$ for motion prediction, the classification head $h_j(\cdot)$ for solving jigsaw puzzles, and
the projection head $h_c(\cdot)$ for contrastive learning are $1$ FC layer ($dim = 60$). The classifier $C(\cdot)$ includes a $1$-layer unidirectional GRU with $60$ units to
be compatible with the dimensions of the output of the encoder $f(\cdot)$ and an MLP for recognition. All networks are initialized with a random uniform distribution.

To optimize our network, Adam optimizer~\cite{newey1988adaptive} is used and the learning rate declines from $0.01$ to $0.0001$ with $0.1$ decay rate for every 100 iterations.
We train the network on one NVIDIA Titan X GPU with a batch size of $32$ for NW-UCLA and $128$ for NTU RGB+D, PKUMMD datasets, respectively.

\subsection{Evaluation and Comparison}
In this section, we explore whether the representations learned by our multi-task self-supervised model (MS$^2$L) are meaningful for action recognition. To give a comprehensive and thorough evaluation, we conduct experiments under different settings, including unsupervised, semi-supervised and fully supervised approaches. We also show the comparison results with other state-of-the-art methods, respectively.

\noindent\textbf{Unsupervised Approaches.}
In the unsupervised setting, the feature extractor, \emph{i.e.}, the encoder $f(\cdot)$, is independently trained with some pretext tasks. Then, the feature representation is evaluated by classifiers. In our experiments, we evaluate feature representations with a linear classifier, which is trained on top of the frozen encoder $f(\cdot)$, and action recognition accuracy is used as a measurement for representation quality. We test the following configurations: 

$\bullet$ \textbf{MS$^2$L Rand-Unsupervised (MS$^2$L Rand-U)}: We only train the linear classifier and freeze the encoder $f(\cdot)$ which is randomly initialized. We regard this configuration as our baseline.

$\bullet$ \textbf{LongT GAN~\cite{zheng2018unsupervised}}: This work designs a conditional skeleton inpainting architecture for learning a fix-dimensional representation with additional adversarial training strategies. Specifically, this model uses the feature of the original data and randomly masked skeleton data to recover the original data. And the trained weights of the encoder $f(\cdot)$ can be used for recognition. We construct the network according to the paper.

$\bullet$ \textbf{MS$^2$L}: It is our full system, where the encoder $f(\cdot)$ is trained by MS$^2$L independently, then we train the linear classifier with encoder $f(\cdot)$ fixed.

In Table~\ref{table:unsupervised}, we show the results of the baseline (\textit{MS$^2$L} Rand-U), the prior work \textit{LongT GAN}, and the proposed \textit{MS$^2$L}. As we can see, our approach achieves better performance over random baseline and \textit{LongT GAN}. This improvement verifies that our methods can force the network to extract more effective features. It is worth noting that the feature dimension from our encoder $f(\cdot)$ is 60 while that from \textit{LongT GAN} is 800. Therefore, we achieve better performance with much more compact feature representations compared to \textit{LongT GAN}.

\begin{table*}
  \caption{Comparison of action recognition results with unsupervised learning approaches.}
  \label{table:unsupervised}
  \begin{tabular}{lcccc}
    \toprule
    Models&NW-UCLA&PKUMMD part I&PKUMMD part II&NTU\\
    \midrule
    MS$^2$L Rand-U & 60.61& 62.80&20.70&41.10\\
    LongT GAN & 74.30& \textcolor{red}{67.70} & 25.95 & 52.14\\
    MS$^2$L (\textbf{Our}) & \textcolor{red}{76.81}& 64.86&\textcolor{red}{27.63}&\textcolor{red}{52.55}\\
  \bottomrule
\end{tabular}
\end{table*}

\begin{comment}
\begin{table*}
  \caption{Comparison of action recognition results with semi-supervised learning approaches.}
  \label{table:semisupervised}
  \begin{tabular}{lcccc}
    \toprule
    Models&NW-UCLA&PKUMMD part I&PKUMMD part II & NTU\\
    \midrule
    $10\%$ \textit{labeled data}:\\
    MS$^2$L Rand-SS & 58.63& 67.95&22.81&62.49\\
    LongT GAN & 59.94& 69.51 & 25.71 & 62.03\\
    MS$^2$L (\textbf{Our})&\textcolor{red}{60.45}& \textcolor{red}{70.30} & \textcolor{red}{26.10}&\textcolor{red}{65.17}\\
    \midrule
    $1\%$ \textit{labeled data}:\\
    MS$^2$L Rand-SS & 17.09& 34.46 &11.79&32.18\\
    LongT GAN & 18.26& 35.78 & 12.37 & \textcolor{red}{35.22}\\
    MS$^2$L Pretrain (\textbf{Our})&21.28& \textcolor{red}{36.42} &\textcolor{red}{13.03}&33.10\\
    MS$^2$L Moving (\textbf{Our})& 19.63 & 35.89 &12.11 & 32.01\\
    MS$^2$L Jointly (\textbf{Our})& \textcolor{red}{22.72}& 36.24 & 12.05 & 34.17\\
  \bottomrule
\end{tabular}
\end{table*}
\end{comment}

\begin{table*}
  \caption{Comparison of action recognition results with semi-supervised learning approaches.}
  \label{table:semisupervised}
  \begin{tabular}{lcccc}
    \toprule
    Models&NW-UCLA&PKUMMD part I&PKUMMD part II & NTU\\
    \midrule
    $1\%$ \textit{labeled data}:\\
    MS$^2$L Rand-SS & 17.09& 34.46 &11.79&32.18\\
    LongT GAN & 18.26& 35.78 & 12.37 & \textcolor{red}{35.22}\\
    MS$^2$L (\textbf{Our})&\textcolor{red}{21.28}& \textcolor{red}{36.42} &\textcolor{red}{13.03}&33.10\\
    \midrule
    $10\%$ \textit{labeled data}:\\
    MS$^2$L Rand-SS & 58.63& 67.95&22.81&62.49\\
    LongT GAN & 59.94& 69.51 & 25.71 & 62.03\\
    MS$^2$L (\textbf{Our})&\textcolor{red}{60.45}& \textcolor{red}{70.30} & \textcolor{red}{26.10}&\textcolor{red}{65.17}\\
  \bottomrule
\end{tabular}
\end{table*}

\noindent\textbf{Semi-Supervised Approaches.} In semi-supervised learning, the training process utilizes both labeled data and unlabeled data. Generally, the encoder $f(\cdot)$ is pretrained with some pretext tasks with unlabeled data, then jointly trained with the classifier $C(\cdot)$ with labeled data. In our experiments, we respectively sample $1\%$ and $10\%$ data randomly from the training set as \textit{labeled data} and regard the rest as \textit{unlabeled data}.

$\bullet$ \textbf{MS$^2$L Rand-Semi supervised (MS$^2$L Rand-SS)}: The encoder $f(\cdot)$ is initialized with random weights. Then we finetune all the weights of the network with \emph{labeled data}. 

$\bullet$ \textbf{LongT GAN~\cite{zheng2018unsupervised}}: To apply this work in semi-supervised learning, we train the weights of GAN with \emph{unlabeled data} and then finetune all the weights with \emph{labeled data}.

$\bullet$ \textbf{MS$^2$L}: We train our full system in the semi-supervised setting. We use both \emph{labeled data}  and \emph {unlabeled data} to independently train the  encoder $f(\cdot)$ with MS$^2$L. Then we train the classifier $C(\cdot)$ and finetune the encoder $f(\cdot)$ with \textit{labeled data} jointly.
%$\bullet$ \textbf{MS$^2$L Pretrain}: We train our full system in the semi-supervised setting. We use both \emph{labeled data}  and \emph {unlabeled data} to independently train the  encoder $f(\cdot)$ with \textit{MS$^2$L}.  Then we train the classifier $C(\cdot)$ and finetune the encoder $f(\cdot)$ with \textit{labeled data} jointly.

%$\bullet$ \textbf{MS$^2$L Moving}: This strategy is to train the network with moving strategy, introduced in Sec.3.2. The encoder $f(\cdot)$ is pretrained with \emph{labeled data}. Then the encoder $f(\cdot)$ and classifier $C(\cdot)$ are trained jointly by switching the pretext tasks and classification task linearly.

%$\bullet$ \textbf{MS$^2$L Jointly}: We train the model from scratch with self-supervised tasks and supervised task at the same time with fixed $\omega$ in Eq.~\ref{equ:joint_train}.

From Table~\ref{table:semisupervised}, we can notice that only with a small subsets of the datasets, our method can always improve the baseline considerably and performs better than \textit{LongT GAN}. 

\noindent\textbf{Supervised Approaches.}
In the supervised setting, the encoder $f(\cdot)$ is pretrained by pretext tasks, and then the encoder $f(\cdot)$ and classifier $C(\cdot)$ are jointly trained with the full training data. We first evaluate our proposed approach in the supervised setting with different training strategies introduced in Sec. 3.2.
The configurations are as follows:

$\bullet$ \textbf{MS$^2$L Rand-Supervised (MS$^2$L Rand-S)}: Our baseline structure initializes the weights of the encoder $f(\cdot)$ randomly and learns them with action labels jointly with the classifier $C(\cdot)$.

$\bullet$ \textbf{MS$^2$L Pretrain}: We initialize the encoder $f(\cdot)$ with the learned weights from self-supervised tasks and then learn the classifier $C(\cdot)$ for action recognition with encoder $f(\cdot)$ fixed.

$\bullet$ \textbf{MS$^2$L Moving}: We train the network with moving strategy, introduced in Sec.3.2. The encoder $f(\cdot)$ is pretrained with pretext tasks. Then we train the encoder $f(\cdot)$ and classifier $C(\cdot)$ jointly by switching the pretext tasks and classification task gradually.

$\bullet$ \textbf{MS$^2$L Jointly}: This strategy requires to train the model with self-supervised tasks and supervised task at the same time. We train the network from scratch with both self-supervised tasks and supervised task with fixed $\omega$ in Eq.~\ref{equ:joint_train}.

The results on the NW-UCLA, NTU and PKUMMD datasets are shown in Table~\ref{table:ucla}, ~\ref{table:pkummd}, ~\ref{table:ntu}, respectively. We improve the performance from $83.86\%$ to $85.32\%$ and $86.75\%$ by
moving pretraining and jointly training on NW-UCLA. And on larger datasets, the performance improves from $83.49\%$ to $84.43\%$ and $85.17\%$ on PKUMMD part I and from $40.97\%$ to $42.57\%$ and $45.70\%$ on PKUMMD part II by moving pretraining and jointly training, respectively. The best performance also improves from $78.44\%$ to $78.56\%$ on NTU dataset.

\begin{table}
  \caption{Comparison of action recognition results with supervised learning approaches on the NW-UCLA dataset.}
  \label{table:ucla}
  \begin{tabular}{lc}
    \toprule
    Models&NW-UCLA\\
    \midrule
    HBRNN-L~\cite{CVPR15HRNN} & 78.50\\
    SK-CNN~\cite{liu2017enhanced} & 86.10\\
    VA-LSTM~\cite{zhang2017view} & 70.71\\
    Denoised-LSTM~\cite{demisse2018pose} & 80.30\\
    \midrule
    MS$^2$L Rand-S & 83.86\\
    MS$^2$L Pretrain (\textbf{Our})&85.26\\
    MS$^2$L Moving (\textbf{Our})&85.32\\
    MS$^2$L Jointly (\textbf{Our})&\textcolor{red}{86.75}\\
  \bottomrule
\end{tabular}
\end{table}

\begin{table}
  \caption{Comparison of action recognition results with supervised learning approaches on the PKUMMD dataset.}
  \label{table:pkummd}
  \begin{tabular}{lcc}
    \toprule
    Models& Part I & Part II\\
    \midrule
    ST-GCN~\cite{yan2018spatial} & 84.07 & 48.20 \\
    VA-LSTM & 84.10 & \textcolor{red}{50.00}\\
    \midrule
    MS$^2$L Rand-S & 83.49&40.97\\
    MS$^2$L Pretrain (\textbf{Our})& 83.46&42.45\\
    MS$^2$L Moving (\textbf{Our})& 84.43&42.57\\
    MS$^2$L Jointly (\textbf{Our})& \textcolor{red}{85.17}&45.70\\
  \bottomrule
\end{tabular}
\end{table}

\begin{table}
  \caption{Comparison of action recognition results with supervised learning approaches on the NTU dataset.}
  \label{table:ntu}
  \begin{tabular}{lc}
    \toprule
    Models&NTU\\
    \midrule
    LSTM~\cite{LSTM1997} & 71.90\\
    BLSTM & 71.40\\
    STA-LSTM~\cite{song2018spatio}&73.40\\
    TPN~\cite{wang2017temporal}&75.30\\
    ST-GCN~\cite{yan2018spatial} &\textcolor{red}{81.50}\\
    \midrule
    MS$^2$L Rand-S &78.44\\
    MS$^2$L Pretrain (\textbf{Our})&78.33\\
    MS$^2$L Moving (\textbf{Our})&78.46\\
    MS$^2$L Jointly (\textbf{Our})&78.56\\
  \bottomrule
\end{tabular}
\end{table}

Compared to the baseline, our self-supervised learning method achieves significant improvement.
Using the moving pretrained strategy helps to make the network change from self-supervised tasks to supervised task gradually
and remember the prior knowledge learned by self-supervised tasks. It is also observed that jointly training achieves more gain than moving pretraining strategy. We explain it as that jointly training can force the network to extract features for different tasks, so the features can be relatively more general and contain richer information. 

Compared with the state-of-the-art, our model performs better on NW-UCLA and PKUMMD part I datasets and can be competitive to the previous methods on PKUMMD part II and NTU datasets.  

\noindent\textbf{Transfer Learning Performance.}
To further evaluate whether the proposed MS$^2$L is able to gain knowledge to related tasks, we investigate the transfer learning performance of our model.

Generally, the representations learned from large scale data more generalizable, as illustrated in \cite{He2019}. 
Therefore, in our experiments,  we regard the NTU and PKUMMD part I as source datasets and PKUMMD part II as the target dataset. We pretrain our model with the source datasets respectively and then fine-tune the whole network on the target dataset. We evaluate transfer learning performance using the accuracy of action recognition on PKUMMD part II and compare the results with those trained with full supervision from PKUMMD part II and \textit{LongT GAN}.
The configurations are as follows:

$\bullet$ \textbf{MS$^2$L Rand-Transfer (MS$^2$L Rand-T)}: We initialize the network randomly and then finetune the whole weights from scratch on PKUMMD part II.

$\bullet$ \textbf{LongT GAN~\cite{zheng2018unsupervised}}: To perform transfer learning with \textit{LongT GAN}, we pretrain the generator on the source dataset and then train the entire network on the target dataset.

$\bullet$ \textbf{MS$^2$L}: The encoder $f(\cdot)$ is trained by self-supervised tasks independently on source datasets, then we train the full system on the target dataset.

Table~\ref{table:transfer} shows the transfer learning results. Our self-supervised model outperforms the supervision baseline, improving the result from $40.97\%$ to $44.14\%$ when pretrained on PKUMMD part I and $45.81\%$ when pretrained on NTU, respectively.

Compared with \textit{LongT GAN}, our model can also show superiority. \textit{LongT GAN} employs adversarial training strategies to reconstruct the whole skeleton data. Therefore, the network is trained to focus more on the details of skeletal joints. The domain gap in detailed skeleton settings of different datasets makes it hard to transfer the knowledge from the source dataset to the target dataset. Our proposed method, however, maps the skeleton data from different datasets to a common feature space with contrastive learning, and then achieve high-level domain knowledge transfer. The results in Table~\ref{table:transfer} illustrate the superiority of our proposed approach.

\begin{table}
  \caption{Comparison of the transfer learning performance.}
  \label{table:transfer}
  \begin{center}
    \begin{tabular}{lc}
      \toprule
      Models&Accuracy on PKUMMD part II\\
      \midrule
      MS$^2$L Rand-T & 40.97 \\
      \midrule
      \textit{PKUMMD part I:}\\
      LongT GAN & 43.61\\
      MS$^2$L (\textbf{Our}) & \textcolor{red}{44.14}\\
      \midrule
      \textit{NTU:}\\
      LongT GAN & 44.83\\
      MS$^2$L (\textbf{Our}) &  \textcolor{red}{45.81}\\
      \bottomrule
    \end{tabular}
  \end{center}
\end{table}

\subsection{Ablation Study}
Next, we conduct ablation experiments to give more analysis of our proposed approach. All the ablation studies are performed on the NW-UCLA dataset. 

\noindent\textbf{Analysis of Self-Supervised Tasks.} In this part, we explore the role that each self-supervised task plays in the learning process. The baseline is training the classifier $C(\cdot)$ independently and the encoder $f(\cdot)$ is with random weights. For evaluating the self-supervised tasks, we  pretrain the encoder $f(\cdot)$ and then finetune the overall network.

\begin{table}
  \caption{Comparison of combinations of self-supervised tasks.}
  \label{table:self}
  \begin{center}
    \begin{tabular}{lcc}
      \toprule
      Method&NW-UCLA\\
      \midrule
      MS$^2$L Rand-S &  83.86\\
      \midrule
      Prediction & 84.88\\
      Jigsaw & 84.11 \\
      Contrastive & 85.47 \\
      \midrule
      Prediction \& Jigsaw & 84.90 \\
      Contrastive \& Jigsaw & 84.82 \\
      Prediction \& Contrastive &  85.71\\
      \midrule
      Prediction \& Jigsaw \& Contrastive & \textcolor{red}{86.75} \\
      \bottomrule
    \end{tabular}
  \end{center}
\end{table}

$\bullet$ \textbf{Motion Prediction:} There are two ways to conduct motion prediction, \emph{i.e.}, temporal motion prediction and spatial motion prediction. For the temporal motion prediction, our model generates
future skeleton data conditioned on the past skeleton data, while the spatial motion prediction reads in the corrupted input sequence, of which a number parts of human key-points are masked and set to be zero, and we predict the masked regions. We pretrain the encoder $f(\cdot)$ with motion prediction and then finetune the whole network jointly. Table~\ref{table:single} shows the results of temporal and spatial motion prediction, respectively. We observe that the spatial masked data reconstruction task does not improve and even hurts their performance. We analyze it may be caused by two reasons. On the one hand, a spatially masked skeleton sequence may lose critical information to infer the unknown skeleton regions. Therefore, it is more difficult to predict the  spatial coordinates. On the other hand, we apply GRU units as our backbone, which explicitly learn temporal patterns but ignore spatial modeling. With a more powerful backbone, such as a graphic model which explicitly models spatial relations, we may boost the performance from spatial motion prediction. We leave it as our future work. In our model, we use temporal motion prediction as one of the self-supervised tasks.

$\bullet$ \textbf{Jigsaw Puzzles:} We explore two different methods to shuffle the skeleton data to perform jigsaw puzzle recognition, \emph{i.e.}, temporal jigsaw puzzle recognition, and spatial jigsaw puzzle recognition. Temporal jigsaw puzzle shuffles the original data temporally and our goal is to predict the correct order of shuffled sequences , while spatial jigsaw puzzle
shuffles the order of key-points which are divided into five parts, namely four limbs and a trunk, and our goal is to recognize each part. Also, the tasks are trained jointly and 
evaluated in classification accuracy. We show the results in Table~\ref{table:self}, utilizing the spatial information harms the performance of the recognition task. It is mainly because the spatial jigsaw is much more difficult to predict since there are more permutations than a temporal jigsaw. In our model, we choose temporal jigsaw puzzle recognition as one of the self-supervised tasks.

$\bullet$ \textbf{Contrastive Learning:} We evaluate different combinations of transformation operators for contrastive learning. In Table~\ref{table:single}, we can observe 
that when we use spatial transformation operators always hurt recognition performance, whereas temporal transformation operators improve performance better. It is illustrated that the features from spatially masked skeletons and spatially shuffled skeletal joints lose much information to establish mapping with the features from original skeletons. 

Table~\ref{table:self} shows the results from a single self-supervised task as well as their combinations.
When applying a single self-supervised task, we can observe that the contrastive learning task always outperforms over the other two tasks. 
Contrastive learning aims to learn a common space between the original and transformed skeleton data. 
Therefore, the network is encouraged to learn more inherent feature representations. 
When two different self-supervised tasks are jointly optimized, they can enhance the performance because of the stronger restriction to the representation space.
And using all the three tasks achieves the best performance. 
We explain it as the features extracted by the tasks jointly keep more aspects of information from the original sequences.
That means the encoder $f(\cdot)$ can extract more general features.

\begin{table}
  \caption{Analysis of self-supervised tasks. (P means Prediction; J means Jigsaw)}
  \label{table:single}
  \begin{center}
    \begin{tabular}{lcc}
      \toprule
      Method&NW-UCLA\\
      \midrule
      MS$^2$L Rand-S &  83.86\\
      \midrule
      Temporal motion prediction&  84.88\\
      Spatial motion prediction & 81.11\\
      \midrule
      Temporal jigsaw puzzle recognition& 84.11\\
      Spatial jigsaw puzzle recognition& 81.62\\
      \midrule
      Contrastive(Spatial P + Temporal J)  & 81.35\\
      Contrastive(Temporal P + Spatial J) &  82.84\\
      Contrastive(Spatial P + Spatial J) & 82.45\\
      Contrastive(Temporal P + Temporal J) & \textcolor{red}{85.47}\\
      \bottomrule
    \end{tabular}
  \end{center}
\end{table}

\begin{figure}[h]
  \centering
  \subfigure[MS$^2$L Pretrain] { 
  \label{fig:a}     
  \includegraphics[width=0.22\textwidth]{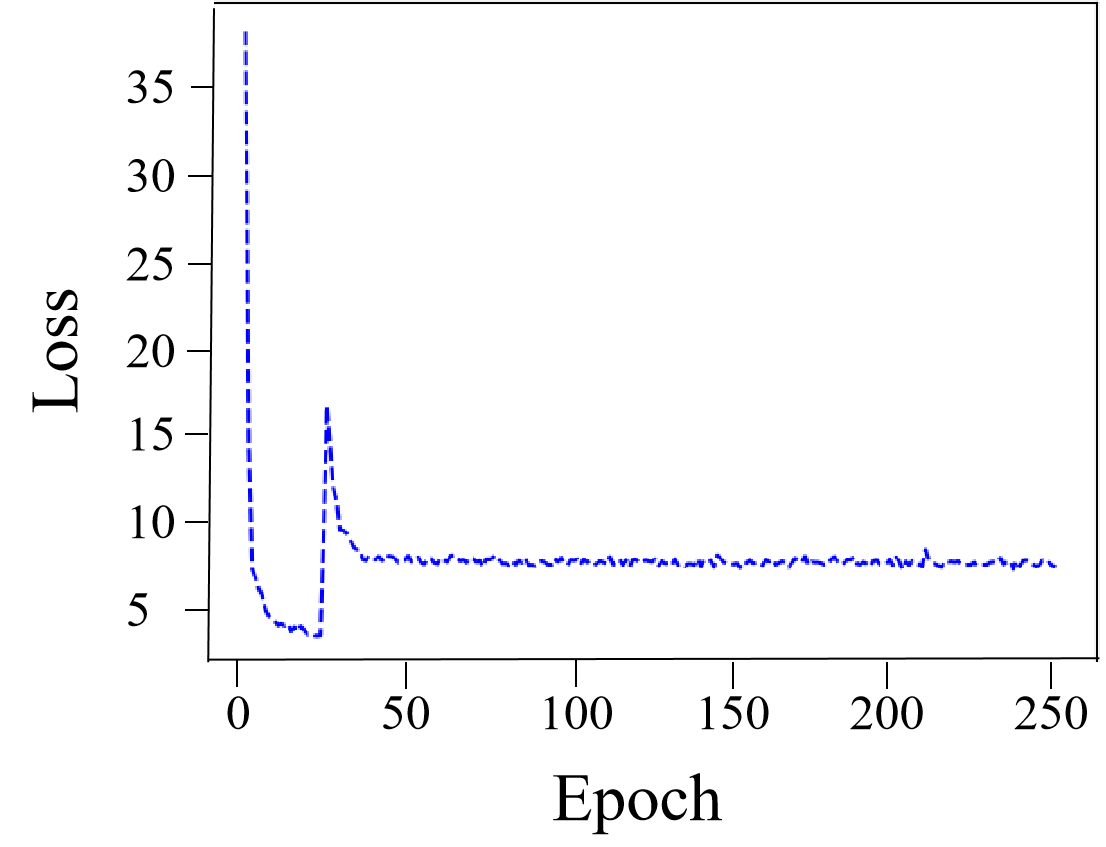}   
  } 
  \subfigure[MS$^2$L Moving] { 
  \label{fig:b}     
  \includegraphics[width=0.22\textwidth]{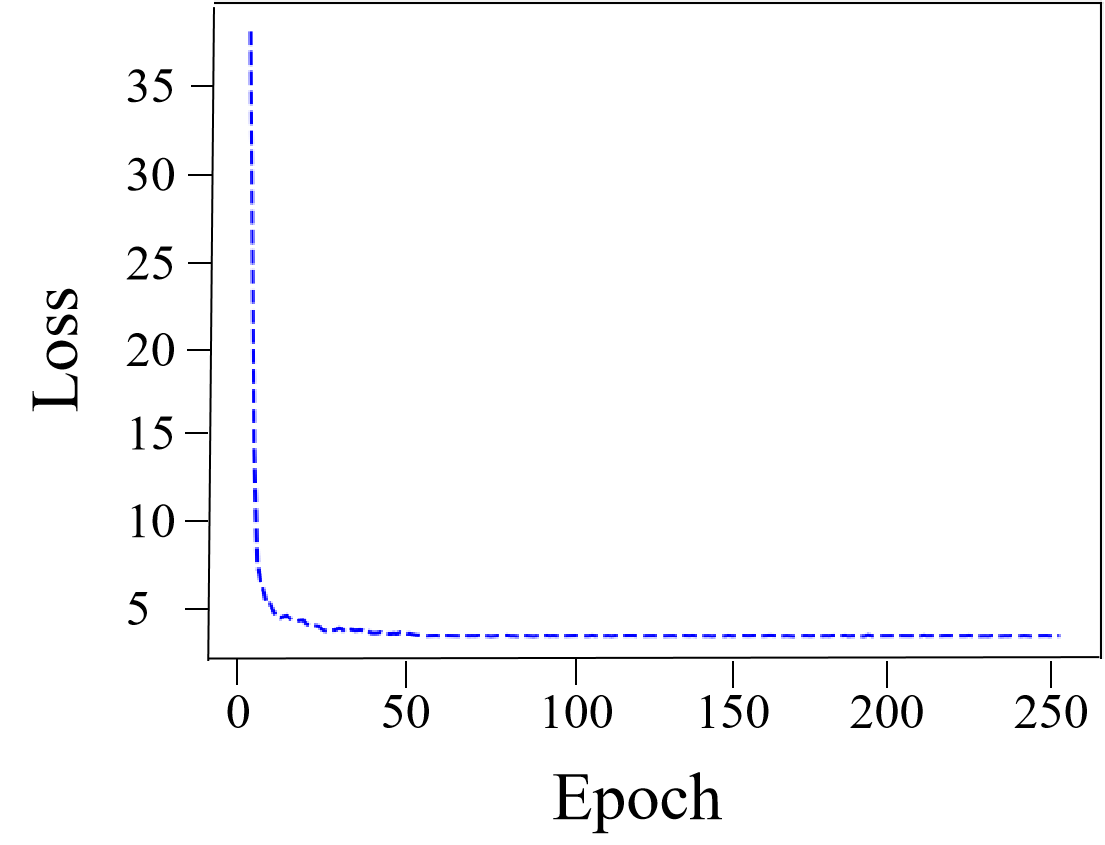}
  } 
  \caption{Loss curves of MS$^2$L Pretrain and MS$^2$L Moving, respectively.}
  \label{figure:moving}
  \vspace{-3mm}
\end{figure}

\noindent\textbf{Training Strategy.} Now we provide some insights into our moving pretraining strategy.
Figure~\ref{figure:moving} shows the sum of all the losses of self-supervised tasks with different training strategies. Figure~\ref{fig:a} shows the losses when we pre-train the self-supervised tasks and fine-tune the overall network for action recognition. We can observe that when we begin to finetune the overall network by jointly training the encoder $f(\cdot)$ and the classifier $C(\cdot)$, the losses of self-supervised learning increase sharply, which may destroy the feature representations learned from the self-supervised tasks. Figure~\ref{fig:b} shows the losses of self-supervised tasks applied by our moving method. The smooth training loss curve illustrates the network is able to learn weights for action recognition while keeping the feature representations learned from the self-supervised tasks. The better results in Tables~\ref{table:ucla}, ~\ref{table:pkummd}, ~\ref{table:ntu} confirm the effectiveness of moving pretraining.

\section{Conclusion}
In this work, we propose a self-supervised learning approach for skeleton-based action recognition. To deal with the overfitting problem of learning skeleton representations from a single reconstruction task, we integrate multiple tasks to learn more general features. We apply motion prediction to model skeleton dynamics and jigsaw puzzle recognition to model temporal patterns, respectively. Besides, contrastive learning is adopted to further regularize the feature space and help learn intrinsic features. With comprehensive and thorough experiments on three datasets, we can show our model is a powerful feature extractor which outperforms the baseline significantly. 

\bibliographystyle{ACM-Reference-Format}
\balance
\bibliography{MS2L_BIB}

\end{document}